\documentclass[runningheads,anonymous=false]{llncs}
\usepackage[T1]{fontenc}
\usepackage{graphicx}
\usepackage[table,xcdraw]{xcolor}
\usepackage{blindtext} 
\usepackage{amsfonts,amsmath,algorithm,algpseudocode,pdfpages}
\usepackage{acronym}

\algrenewcommand\algorithmicrequire{\textbf{Input:}}
\algrenewcommand\algorithmicensure{\textbf{Output:}}
\begin{document}
\title{Building surrogate models using trajectories of agents trained by Reinforcement Learning}
\titlerunning{Building surrogate models with RL agents}

\author{Julen Cestero\inst{1,2}\orcidID{0000-0002-6670-6255} \and
Marco Quartulli\inst{2}\orcidID{0000-0001-5735-2072} \and
Marcello Restelli\inst{1}\orcidID{0000-0002-6322-1076}}
\authorrunning{Cestero et al.}
\institute{Politecnico di Milano, Milano, Italy \\
Vicomtech, San Sebastian, Spain}


\maketitle              
\begin{abstract}
Sample efficiency in the face of computationally expensive simulations is a common concern in surrogate modeling.  Current strategies to minimize the number of samples needed are not as effective in simulated environments with wide state spaces. As a response to this challenge, we propose a novel method to efficiently sample simulated deterministic environments by using policies trained by Reinforcement Learning. We provide an extensive analysis of these surrogate-building strategies with respect to Latin-Hypercube sampling or Active Learning and Kriging, cross-validating performances with all sampled datasets. The analysis shows that a mixed dataset that includes samples acquired by random agents, expert agents, and agents trained to explore the regions of maximum entropy of the state transition distribution provides the best scores through all datasets, which is crucial for a meaningful state space representation. We conclude that the proposed method improves the state-of-the-art and clears the path to enable the application of surrogate-aided Reinforcement Learning policy optimization strategies on complex simulators.

\end{abstract}
\begin{acronym}[MPC] 
\acro{AL}{Active Learning}
\acro{Random}{Random sampling}
\acro{Sobol}{Sobol sampling}
\acro{LHS}{Latin-Hypercube sampling}
\acro{RA}{Random agent}
\acro{EA}{Expert agent}
\acro{MEA}{Maximum Entropy agent}
\acro{MA}{Mixed agent}
\acro{MPA}{Mixed (Partially) agent}
\acro{PA}{Partial agents}

\end{acronym}
\section{Introduction}
Over the past few years, the gradual integration of Machine Learning techniques into a variety of processes has enhanced their efficiency and effectiveness. A widely used approach to optimize these processes involves the development of simulators that allow algorithmic solutions to iterate with a more efficient and safe version of the real elements integrated into the process. These simulators have quickly evolved into invaluable tools for testing and refining control strategies, thus reducing operational risks and minimizing downtime. Despite their significance, several methods for optimizing these processes rely on expensive data acquisition or intricate micro-optimizations to accommodate the training of ML models \cite{asher_review_2015,broadWaterDistributionSystem2005,caoModelFreeVoltageRegulation2020}.

In response to this challenge, surrogate models have emerged as a compelling solution. These models are used when the actual relationship between the input and output values of simulators are computationally expensive to evaluate and, therefore, can be used as substitute models for the actual simulators. 

This paper proposes a new type of surrogate models, based on Reinforcement Learning (RL) agents, which are focused on exploring realistic state transitions in simulated environments. In other words, the models trained with these datasets present high accuracy around the subset of states that appear in typical use-trajectories of the simulator rather than more stable (but lower) accuracy on the whole state space. The datasets collected for training the surrogate models are acquired by collecting trajectories using different control policies, such as a random policy, a policy trained to optimize the environment's objective, which we call `expert policy', or a policy that aims to maximize the entropy of the state-visit distribution \cite{muttiTaskAgnosticExplorationPolicy2021}. We empirically prove that this maximum entropy policy provides critical information to generalize the dynamics of the simulated environment.

In this work, we compare the performance of differently trained surrogate models. On the one hand, there are more classical methods for acquiring the datasets, sampling the whole state space or using Design of Experiments (DoE) methods \cite{eriksson2000design}, as well as the well-known Kriging method \cite{cressieOriginsKriging1990}  for building surrogates. On the other hand, there are the aforementioned novel methods based on RL exploration, which we call `agent-based' sampling methods. 

To evaluate the effectiveness of this research, we studied the proposed methodology using different environments (commonly used in RL) of increasing complexity, such as CartPole, MountainCar, HalfCheetah and Ant.

Our main contributions are two: to propose a new method for constructing surrogate models, taking inspiration from Markov Decision Processes (MDPs) to make it possible to simulate realistic trajectories over the simulated environments (agent-based sampling methods) which outperforms the state-of-the-art methods; and to propose a novel way to sample the environments, acquiring the most information of the state transitions, using a mix of agents, including a task-agnostic agent with the aim of maximizing the entropy of the state visitation distribution, which contributes critically to a correct environment representation.

\subsection{Previous work}

Throughout the years, multiple techniques have been proposed to build surrogate models from simulators. Taking into account data-driven methods \cite{asher_review_2015}, a common strategy to acquire and compose a surrogate model is through Kriging and Gaussian processes \cite{cressieOriginsKriging1990}. 
Many authors propose surrogate construction methods using kriging and Gaussian processes. For example, da Costa et al. \cite{dacostapauloSurrogateModelHVAC2023} suggest a combined method of Active Learning with kriging and Gaussian processes applied to energy optimization and Demand Response. This yields highly accurate surrogate models for multidimensional simulators. In the domain of mechanical simulators, Gaspar et al. \cite{gasparAssessmentEfficiencyKriging2014} evaluate the efficiency of kriging models as surrogates for structural reliability analysis and evaluate the accuracy of the models by comparing the distribution of the predicted data with the real data. Furthermore, Dubourg et al. \cite{dubourgReliabilitybasedDesignOptimization2011} propose a method to solve more abstract problems, specifically reliability-based design optimization problems, which mixes DoE techniques with Kriging metamodeling to improve optimization speed using classical gradient-based optimization algorithms. Similarly, other authors \cite{moustaphaQuantilebasedOptimizationUncertainties2016,xingGlobalOptimizationStrategy2020} propose similar techniques for solving the same problem. 

However, other authors, such as Zhou \cite{zhouEnhancedKrigingSurrogate2020}, point out a limitation of Kriging, stating its inefficiency for high-dimensional simulators or simulators with discontinuities in the target curve. This limitation, coupled with recent advancements in Machine Learning, prompts several authors to suggest alternatives to the classical Kriging. As mentioned by McBride \cite{mcbrideOverviewSurrogateModeling2019}, various surrogate model definitions exist, such as polynomials explored by Forrester \cite{forresterRecentAdvancesSurrogatebased2009} or Simpson et al. \cite{simpsonKrigingModelsGlobal2001}, which consist on using classic polynomial response surfaces that adjust to the simulator's curves; Artificial Neural Networks (ANN) studied by Chollet et al. \cite{cholletDeepLearningPython2021}, which are generalized approximators capable of adapting to many different surfaces; and Radial Basis Functions observed in Fang \cite{fangGlobalResponseApproximation2006} or Wang et al. \cite{wangNovelFeasibilityAnalysis2017}. Finally, Razavi et al. \cite{razaviReviewSurrogateModeling2012} present alternative methods, such as employing typical Machine Learning regressors like Support Vector Machines (SVM) \cite{cherkasskyPracticalSelectionSVM2004} or Metamodeling \cite{broadWaterDistributionSystem2005}.
Our work tries to improve on the limitations above.

\subsection{Related work}

Surrogate building techniques have in common that they all need a sampling strategy of the state space of the different simulated environments. However, although these methods may provide a good average score across the whole space, some simulated environments have a large number of dimensions, which makes a balanced and accurate representation of the state space challenging. Our approach consists of building the best surrogate possible within realistic use trajectories.

Giraldo-Pérez et al. \cite{giraldo-perezReinforcementLearningBased2023} propose a method for building the surrogate of a refrigerator using realistic usage patterns, building a representative dataset of usage data as training data. Sieusahai \cite{sieusahaiExplainingDeepReinforcement2021} proposes a method to make Atari game states interpretable through surrogate models, detecting game regions based on color and pixel count, generating features more easily interpretable by humans. Bui et al. \cite{buiDeepNeuralNetworkBased2022} present a method similar to ours, training surrogate models using a dataset from a simulated power converter and achieving results close to near-perfect precision. Similarly, Cao et al. \cite{caoModelFreeVoltageRegulation2020} also proposed to train a surrogate with transitions in power injections and voltage fluctuations in a voltage regulation system (Active Distribution Networks).

Another approach to solving this problem is to acquire samples using trajectories. Lüthen et al. \cite{luthenSpectralSurrogateModel2023}  propose a method for creating surrogate models using spectral methods and statistical inference. Other methods to efficiently explore state space have been provided by Mutti et al. \cite{muttiTaskAgnosticExplorationPolicy2021}  in Reinforcement Learning environments by creating an algorithm for training an agent that maximizes the entropy of the space variables and, thus, maximizing the exploration using trajectories.

In this work, we explore diverse methods for constructing surrogate models, such as Kriging with Active Learning (as recommended by \cite{dacostapauloSurrogateModelHVAC2023}) or different ML techniques (regressors and ANN models) applied to datasets obtained through various samplings (\ac{LHS}, \ac{Sobol}, \ac {Random}). We introduce a novel approach for constructing surrogate models for using RL environments as an inspiration for creating agent-based samplings, incorporating random policies, expert-based heuristics, and policies maximizing exploration entropy \cite{muttiTaskAgnosticExplorationPolicy2021}.

\section{Methods}\label{sec:methods}
\subsection{MDP Surrogate definition}


We define our simulated environments as Markov Decision Processes (MDPs), because we aim to predict the state transitions. For that, we define our surrogate model as a function $f$, which generates a transition from state $s_t \in \mathcal{S} \subset \mathbb{R}^n$ guided by the action taken by a RL agent $a_t \in \mathcal{A}$, where $\mathcal{A}$ can be $\mathbb{R}^n$ for continuous actions or $\mathbb{Z}^n$ for discrete actions.  The definition of the surrogate simulator, and hence the transitions, is $f: (s_t, a_t) \mapsto (s_{t+1})$. Note that this definition only takes into account deterministic MDPs, on which this work is focused.

\subsection{Modeling techniques}
When discussing surrogate models, two common options are design optimization or design space approximation (emulation) \cite{williamsSurrogateModelSelection2019}. The first one aims to build a good surrogate model around an optimal point of the simulator, therefore, finding this optimal point. The second one seeks to build a surrogate model that is good in the whole space. We focus on the latter.


We employ different modeling techniques based on a quantitative analysis performed by design exploration and AutoML tools \cite{pandalaShankarpandalaLazypredict2024,PyCaret}. For this work, we used three different types of regression models: XGBoost, which is the best model according to the AutoML tools for our data; Artificial Neural Networks (ANN); and Gaussian Process along with Active Learning (Kriging method) as a baseline, using its feature of returning the prediction uncertainty along all the space.

\subsection{Dataset generation methods} \label{sec:dgm}

Generating training datasets is a critical step in constructing surrogate models. One typical method is Kriging \cite{mai_improved_2022}, though various other approaches exist. In this work, we focus on studying the effectiveness of building surrogate models based on the XGBoost predictor and ANNs alongside the more classic Kriging. Additionally, we investigate the effectiveness of using different sampling methods to construct datasets that represent MDPs of environments. We classify the sampling methods into two categories: generative methods and agent-based methods.

Generative methods sample the state space to calculate transitions, while agent-based methods simulate the trajectories of a given agent in the MDP model of the simulator. Both methods have advantages and disadvantages, suitable for different simulator natures and subrogation goals. Generative methods are useful for covering a broader spectrum of the space, but may lack density in specific trajectories. Agent-based methods compensate for this by providing more natural transitions in the dataset.

\subsubsection{Generative methods}

In this work, the generative spacial sampling methods studied are Latin-Hypercube sampling (\ac{LHS}) \cite{heltonLatinHypercubeSampling2003}, \ac{Sobol} sampling \cite{burhenne2011sampling}, \ac{Random} sampling, and we add to this category the Kriging method, for which we use a mix of Gaussian Process regressor and \ac{AL}, although this method uses a different procedure from the others and is covered in \cite{dacostapauloSurrogateModelHVAC2023}. The dataset generation process with these generative methods is as follows: 

\begin{enumerate}
  \item Define $f: (s_t, a_t) \mapsto (s_{t+1})$ as the simulator for MDP transitions. 
  \item Let $\mathcal{O}$ be the entire sample space where state variables $s_t$ are defined. We obtain, at least, one subset $\mathcal{S}_i \subset \mathcal{O}$ using a generative method (\ac{LHS}, \ac{Sobol}, \ac{Random}). $\mathcal{S}_i$ has the following characteristics:
  \begin{enumerate}
    \item $|\mathcal{S}_i| = k$: $\mathcal{S}_i = \{s_1, s_2, ..., s_k\}, \quad s_j \in \mathcal{S} (\subset \mathbb{R}^n)$
    \item $(\forall j, \forall l): \quad  -\infty < s_{\text{min},l} < s_{j,l} < s_{\text{max},l} < \infty$
  \end{enumerate}
  \item Generate transitions $s_{t+1} \in \mathcal{O}, s_{t+1} \notin \mathcal{S}_i$,  by simulating each sample.
\end{enumerate}

This process is applicable to continuous or discrete action variables. For continuous actions, actions are added as part of state variables, and therefore its definition changes to $p_t = (s_t, a_t) \in \mathcal{P}, \quad p_t \in \mathbb{R}^{n+m}, \quad \mathcal{P} \subset (\mathcal{O} \times \mathcal{A})$. If the actions are discrete, instead of combining the action space with the state space, the samples $\mathcal{S}$ are obtained by sampling from $\mathcal{O}$ with all the actions in $\mathcal{A}$.

Having obtained all samples $s_t \in \mathcal{S}_i$, the next step involves simulating all samples along with a combination of actions to obtain state transitions $s_{t+1} \in \mathcal{O}, s_{t+1} \notin \mathcal{S}_i$. A wrapper is used to convert the simulator into an RL environment to simulate each transition, storing the transition states. Therefore, since the samples are independent of each other, it becomes necessary to reset the internal partial states of the simulator in each simulation to prevent data carryover between samples.



\subsubsection{Agent-based methods}

Agent-based methods acquire data in a manner similar to how a RL algorithm collects data through rollouts. The requirements for these methods are akin to generative methods, with the distinction that prior policies are needed to explore the state space $\mathcal{O}$. Instead of randomly sampling the entire $\mathcal{O}$, these methods behave like RL agents and run the simulation from start to finish, guided by policies. The policies used in this work include a Random policy $\pi_r$; a policy referred to as the Expert policy $\pi_e$, which is essentially a policy trained in the original environment with the aim of optimizing the reward function; and an entropy-maximizing policy $\pi_m$ \cite{muttiTaskAgnosticExplorationPolicy2021}, with the objective of maximizing state exploration. The agents resulting from these policies in this work are called \ac{RA}, \ac{EA}, and \ac{MEA}. 

The process of sample acquisition consists of generating trajectories obtained from policy rollouts. Initially, the simulator must be introduced into a wrapper that emulates the functionality of an RL environment, i.e., operating as an MDP and returning the variables $s_{t+1}, r_t,$ and $d_t$, representing the next observation of the environment, the obtained reward and a flag describing if the agent reached a terminal state. Subsequently, the environment is initialized to a random state by a reset, and a loop is executed, simulating steps by selecting actions $a_t \sim \pi(\cdot | s), \quad a_t \in \mathcal{A}$. The distinction between different methods directly influences action selection. The Random policy is the simplest, randomly choosing at each step one of the available actions with equal probability for all actions. The Expert policy is heuristic-based, akin to a policy generated by an expert operator or one seeking the simulator's optimality goal without being considered the actual optimal policy. An example of this policy is using an agent pre-trained in the simulator through RL. Finally, an entropy-maximizing policy \ac{MEA}, which considers an importance-weighted k-NN estimator of Shannon differential information measures of the state-action space  with proven quality guarantees \cite{muttiTaskAgnosticExplorationPolicy2021}. The training sample collection process is represented in Algorithm \ref{alg:agent_sampling}

\begin{algorithm}
    \caption{Dataset construction algorithm for agent-based methods}
    \begin{algorithmic}
        \Require number of samples $k$, simulator $f$, wrapper constructor $W$
        \State Initialize $w = W(f)$ and dataset $\mathcal{D}$
        \State $t = 0$
        \State $i = 0$
        \State Reset $w$ environment
        \State Get initial random state $s_0$ of the episode
        \While{$i < k$}
            \State $a_t =$ Get action from policy $\pi(s_t)$
            \State Compute a simulation step:
            \State\hspace{\algorithmicindent} $s_{t+1} = w(s_t, a_t)$
            \State $\mathcal{D}_i \gets s_t, a_t, s_{t+1}$
            \State $t \gets t + 1$
            \State $i \gets i + 1$
            \State $s_{t} \gets s_{t+1}$ 
            \If{Arrived to terminal state}
                \State Reset $t \gets 0$ 
                \State Reset $w$ environment
                \State Get initial random state $s_0$ of the episode
            \EndIf
        \EndWhile
        \Ensure dataset $\mathcal{D}$
    \end{algorithmic}
    \label{alg:agent_sampling}
\end{algorithm}

\section{Results}
\subsection{Training conditions}
The results of this study have been obtained by training three types of regressors: XGBoost, GaussianProcess, and ANNs. These models have been trained using datasets with $100\,000$ samples obtained by the methods described in Section \ref{sec:methods} with an Intel Core i5-9400F CPU for approximately 10 hours. The specific details of each algorithm are provided in Appendix \ref{app:algo}. 

The loss function used in the training process is the Mean Squared Error (MSE) for each model. However, to evaluate the performance of the trained models, the coefficient of determination metric ($R^2$) has been used. 
 
The datasets used for this study have been obtained by running the implementations described in Section \ref{sec:dgm}: \ac{LHS}, \ac{Random} and \ac{Sobol} sampling as generative methods, and \ac{EA}, \ac{RA} and \ac{MEA} as agent-based methods. We also added three other types of datasets obtained with agent-based methods:  \ac{MA} dataset, which is a dataset obtained by mixing the \ac{EA}, \ac{RA} and \ac{MEA} datasets;  a partially mixed dataset,  \ac{MPA}, which is obtained by mixing only the \ac{EA} and \ac{RA} datasets (thus, a dataset that mixes all the agents without the \ac{MEA} dataset); and some Partial\ac{PA} for some of the cases, which are datasets obtained by partially trained expert policies, used to obtain datasets with non-optimal policies different from the \ac{MEA} and the \ac{RA}.

\subsection{Dataset analysis}

\begin{figure}[t]
   \centering
   \includegraphics [width=1\linewidth]{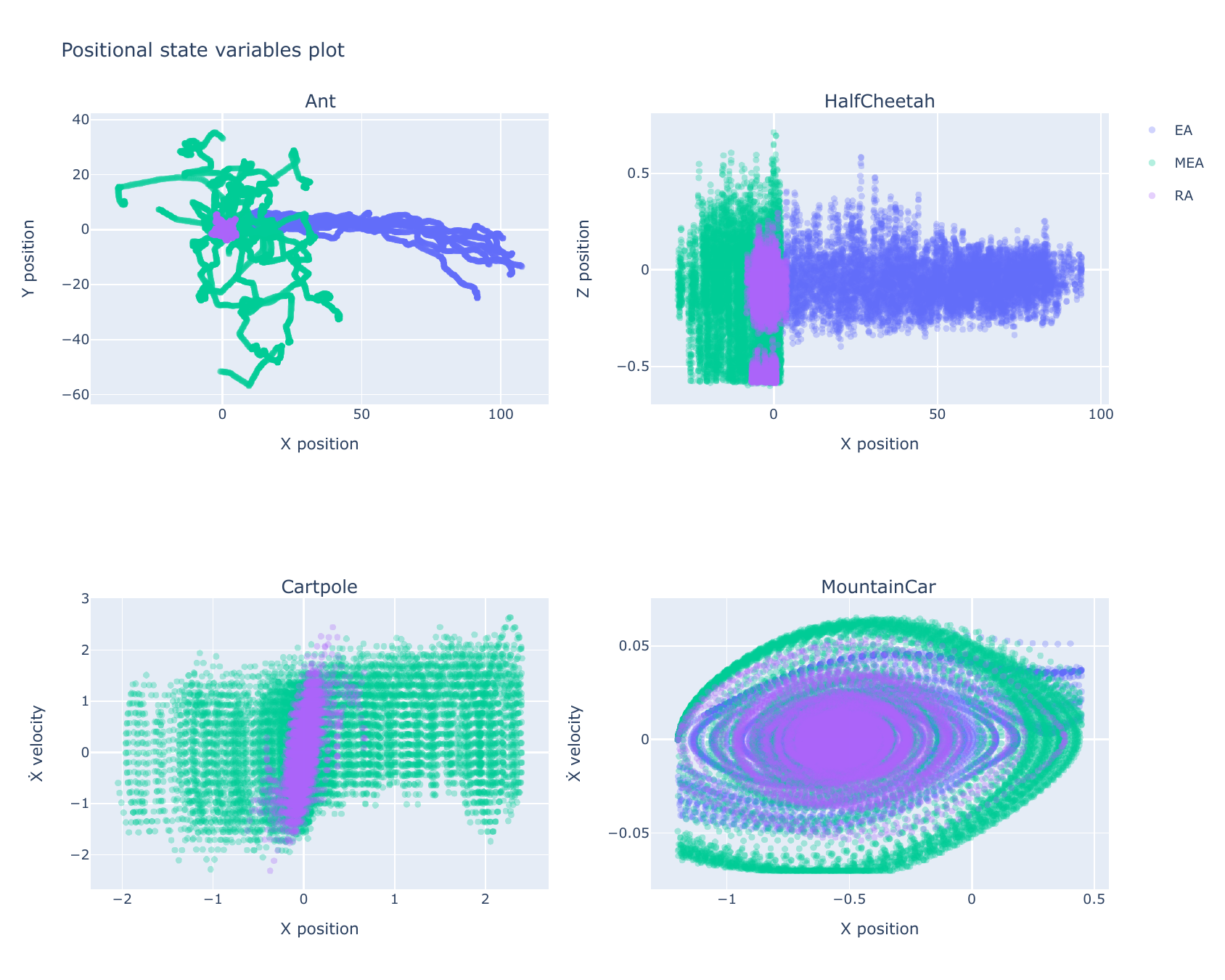}
   \caption{Representation of the first two variables from the environments using different agent-based samplers. For the Mujoco environments, these are positions, while for the simpler environments, the position X and the velocity are shown.}
   \label{fig:XY}
\end{figure}

In this section, we analyze the distributions of the datasets. We assume that the generative samplers take samples uniformly over all the state space. In the Mujoco case, these samples are bounded between $[-10, 10]$;  otherwise, the upper and lower bounds were $(-\infty, \infty)$. The agents do not sample all the state space equally, since some states are hardly reachable and inconsistent to explore, thus, the information gained from these states may not relevant. Therefore, to get a sense of what is the region of the state space that these samplers explore, we consider Figure \ref{fig:XY}. 

This figure depicts the first two variables of the different environments. For Ant, these two positions are the coordinates X and Y of the tridimensional space. For HalfCheetah, these variables are the X position and the height Z. For Cartpole and MountainCar, we depict their X position and their velocity. As it can be seen, the trajectories shown in the figure match what one intuitively may imagine about how each agent may behave. In the Ant case, the \ac{EA} goes to the furthest right side, while the \ac{MEA} explores a large region of the space, balanced between all the variables, and the \ac{RA} explores only the area around the initial state. This repeats throughout the environments: \ac{MEA} covers a larger section of the space, \ac{RA} is restricted to the area around the initial state, and \ac{EA} moves precisely to the original environment's location where the maximum reward is acquired. As a remark, the \ac{EA} is hardly represented in the Cartpole environment because in this particular case, we can expect that this agent would make the minimum movements to balance the pole; thus, the exploration of this agent is smaller than the \ac{RA} and this result is coherent with the results shown in further sections.

\begin{figure}[h]
   \centering
   \includegraphics [width=1\linewidth]{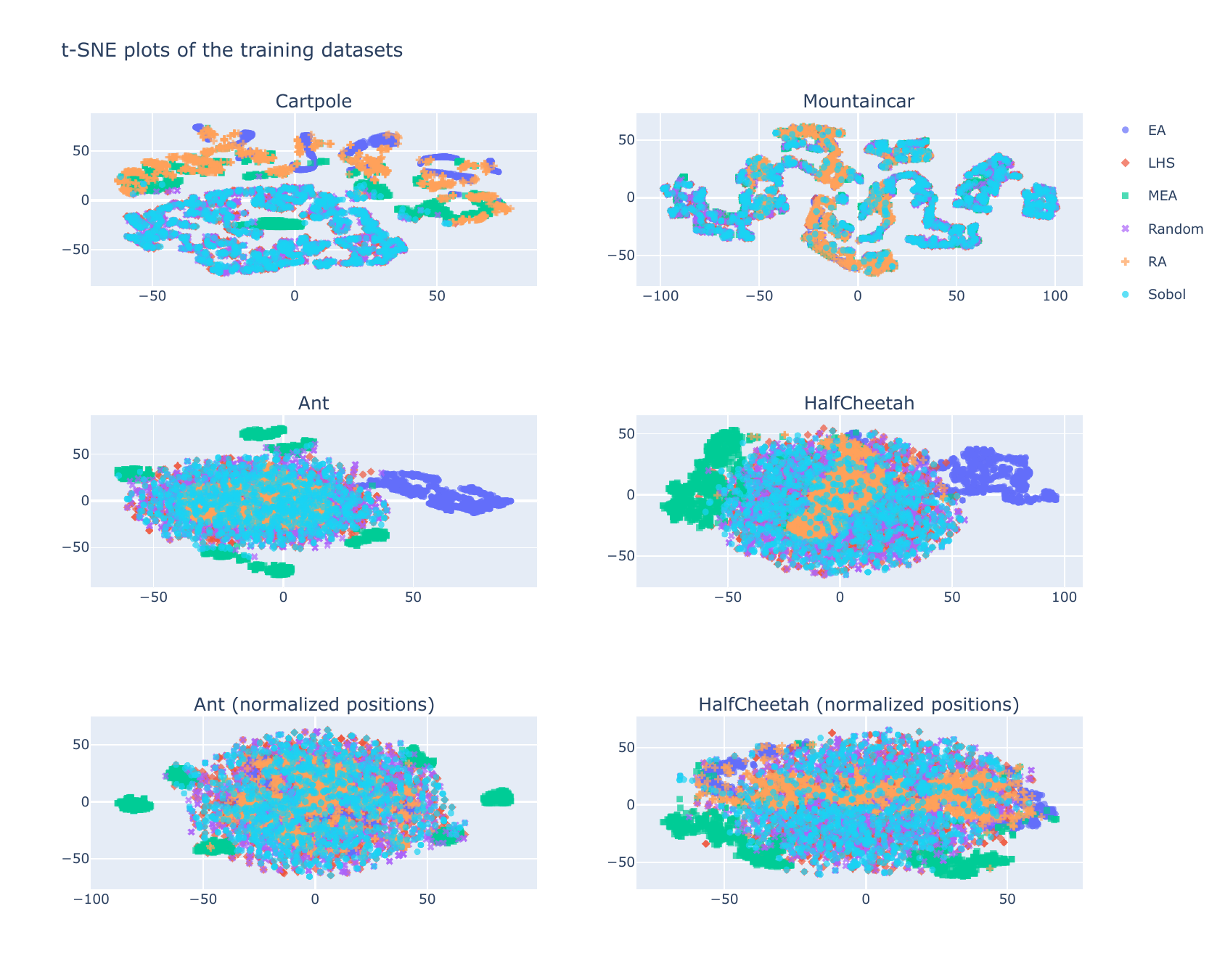}
   \caption{t-SNE map of the manifolds of the original trajectories. Generative samplers are confined to similar areas, while the agent-based samplers are distributed in separated areas.}
   \label{fig:tsne}
\end{figure}

Figure \ref{fig:tsne}  presents the trajectories in a t-SNE map of the manifolds the original trajectories live in \cite{van2008visualizing}. This plot gives us a sense of how the data is distributed throughout all the variables in a two-dimensional projection. In general, all the generative samplers are confined to similar areas, whereas the agent-based samplers occupy different areas. In the case of the Mujoco environments, the \ac{EA} and \ac{MEA} datasets are imbalanced through the X (and Y in the Ant case) positions, and to reduce this effect, we show below the distribution with normalized positions. It can be seen that the \ac{MEA} explores some regions unreachable by other samplers, while the \ac{RA} is highly correlated with the generative samplers in the Mujoco case. For the Cartpole and MountainCar cases, the agents explore regions different from the generative samplers because some regions are unreachable by working agents (surpassing terminal states).

\subsection{Surrogate training results}
In this section, we introduce the training results of the differently trained surrogate models.


\begin{table}[!ht]
    \centering
    \setlength{\tabcolsep}{10pt}
    \caption{Aggregated $R^2$ comparison between the different groups of methods used for training the surrogate models. For simpler environments, classic methods are slightly more efficient, while for the Mujoco environments, the agent-based methods show higher and more robust scores. Best samples in bold as per t-test with $p=0.001$}
    \begin{tabular}{llll}
    
        \textbf{Environment} & \textbf{Surrogate} & \textbf{mean} & \textbf{std} \\ \hline
        & Kriging & 0.807 & 0.292 \\ 
        \textbf{Cartpole} & \textbf{Generative samplers} & \textbf{0.96} & \textbf{0.062}\\ 
        & Agent-based samplers & 0.73 & 0.303 \\ \hline
        & Kriging & 0.919 & 0.136 \\ 
        \textbf{MountainCar} & \textbf{Generative samplers} & \textbf{0.999} & \textbf{0.001} \\
        & Agent-based samplers & 0.988 & 0.016 \\ \hline
        & Kriging & -6e3 &  8.4e3 \\ 
        \textbf{Ant} & Generative samplers & -4.873 &  5.394 \\
        & \textbf{Agent-based samplers} & \textbf{0.613} & \textbf{0.399} \\ \hline
        & Kriging & -3.8e4 &  1.1e6 \\ 
        \textbf{HalfCheetah} & Generative samplers & -19.85 & 32.64 \\ 
        & \textbf{Agent-based samplers} &\textbf{ 0.549 }& \textbf{0.414} \\         
\\ 
    \end{tabular}
    \label{tab:top_avg_general}
\end{table}
We show a general overview of the results of this work in Table \ref{tab:top_avg_general}. This table depicts the averaged results of using the different sampling methods proposed in this work for each studied environment, aggregated among agent-based samplers or generative samplers. We also add kriging in a separate view for the sake of clearer comparison. It can be seen that the generative space samplers have better scores among the simpler environments since Cartpole and MountainCar only have 4 and 2 state variables, bounded tightly around 0, while agent-based samplers are the only ones that work in a fairly robust way in the more complex Mujoco environments. The Kriging method, using \ac{AL}, on the other hand, shows similar results to the generative samplers, but this method fails in complex environments. Therefore, we can assume from these experiments that agent-based samplers can be used both in simple and complex environments.
\begin{figure}[!h]
   \centering
   \includegraphics[width=1\linewidth]{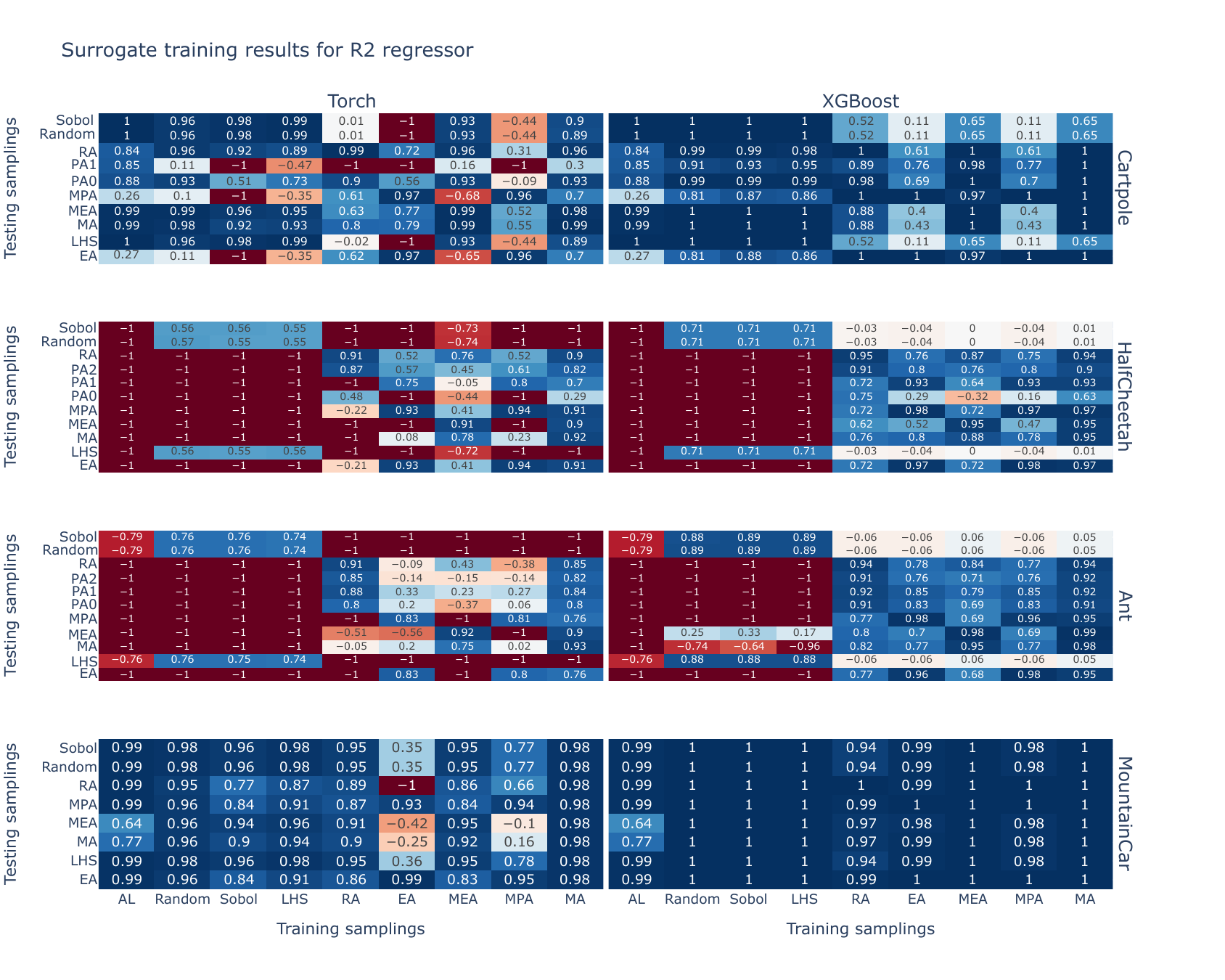}
   \caption{Heatmap with raw $R^2$ cross-validated results of the surrogates trained for all the environments bounded by $[-1, 1]$. These results show the score of the surrogates trained in the datasets obtained by the different methods (X-axis) and are validated against all the testing datasets (Y-axis). The results show a division between the Agent-based and the space sampling methods, while \ac{MA} shows the overall best results over all the datasets.}
   \label{fig:raw_ant}
\end{figure}
Figure \ref{fig:raw_ant} shows the raw $R^2$ scores of cross-validating all the surrogates according to the sampling method used for training them against all the datasets obtained by these sampling methods. Note that in some cases, the scores are bounded to $-1$ for easier reading of the results. We observe the following facts:
\begin{enumerate}
    \item There is a significant difference between agent-based sampling methods and spatial sampling methods. The latter are decent for generalizing broad behaviors within a large space, but they struggle with predicting specific trajectories requiring detail, as opposed to the agent-based methods. These methods, on the other hand, perform well with datasets that contain realistic datasets but have difficulty generalizing to the overall space. However, \ac{MA} and \ac{MEA} clearly outperform other surrogates in unexplored regions of the state space due to the exploratory nature of the \ac{MEA}, which shows the least bad results against the overall space sampling datasets.
    \item XGBoost significantly outperforms ANNs, with better $R^2$ scores and generalization of training.
    \item The \ac{MA} surrogate, combining experience from all agents, performs best overall, while \ac{MPA}, without MEA's dataset within, does not show a significant improvement over \ac{RA}  or \ac{EA}.
\end{enumerate}

\begin{table}[h]
\setlength{\tabcolsep}{7pt}
\centering
\caption{Averaged $R^2$ results over all the testing samplings. Agent-based methods show better scores for the Mujoco environments and have similar performance for narrow-state space environments.}
\begin{tabular}{ll|ll|ll|ll}
\multicolumn{2}{c|}{\textbf{MountainCar}}                  & \multicolumn{2}{c|}{\textbf{Cartpole}}                     & \multicolumn{2}{c|}{\textbf{Ant}}                                    & \multicolumn{2}{c}{\textbf{HalfCheetah}}                             \\ \hline
Sobol                      & 0.9995                        & Sobol                      & 0.9808                        &  \textbf{MA}     & \textbf{0.5726}        &  \textbf{MA}     & \textbf{0.5487}        \\
Random                     & 0.9995                        & LHS                        & 0.9774                        & \textbf{MEA}    & \textbf{0.5181}        & \textbf{MEA}    & \textbf{0.4885}        \\
LHS                        & 0.9995                        & Random                     & 0.9709                        & RA                            & 0.4473                               & RA                            & 0.4245                               \\
 \textbf{MEA} & \textbf{0.9987} & AL                         & 0.8687                        & MPA                           & 0.4335                               & EA                            & 0.4172                               \\
\textbf{MA}  & \textbf{0.9987} & \textbf{MA}  & \textbf{0.8501} & EA                            & 0.4319                               & MPA                           & 0.4108                               \\
EA                         & 0.9871                        & \textbf{MEA} &  \textbf{0.8449} & Random &  $<-1$& Sobol  &  $<-1$ \\
MPA                        & 0.9867                        & RA                         & 0.7507                        &  Sobol  &  $<-1$&  Random&  $<-1$ \\
RA                         & 0.9632                        & EA                         & 0.3981                        & LHS    &  $<-1$& LHS    & $<-1$ \\
AL                         & 0.9085                        & MPA                        & 0.3973                        &  AL     & $<-1$ &  AL     &$<-1$
\end{tabular}

\label{tab:avg_results}
\end{table}


To facilitate understanding, we have averaged the heatmap results in Table \ref{tab:avg_results}. These results display the average $R^2$ metric for all testing datasets.  Agent-based models excel in complex environments with numerous state variables, while traditional methods perform better in lower-dimensional environments, although the difference is minimal.

\begin{figure}[h]
   \centering
   \includegraphics [width=1\linewidth]{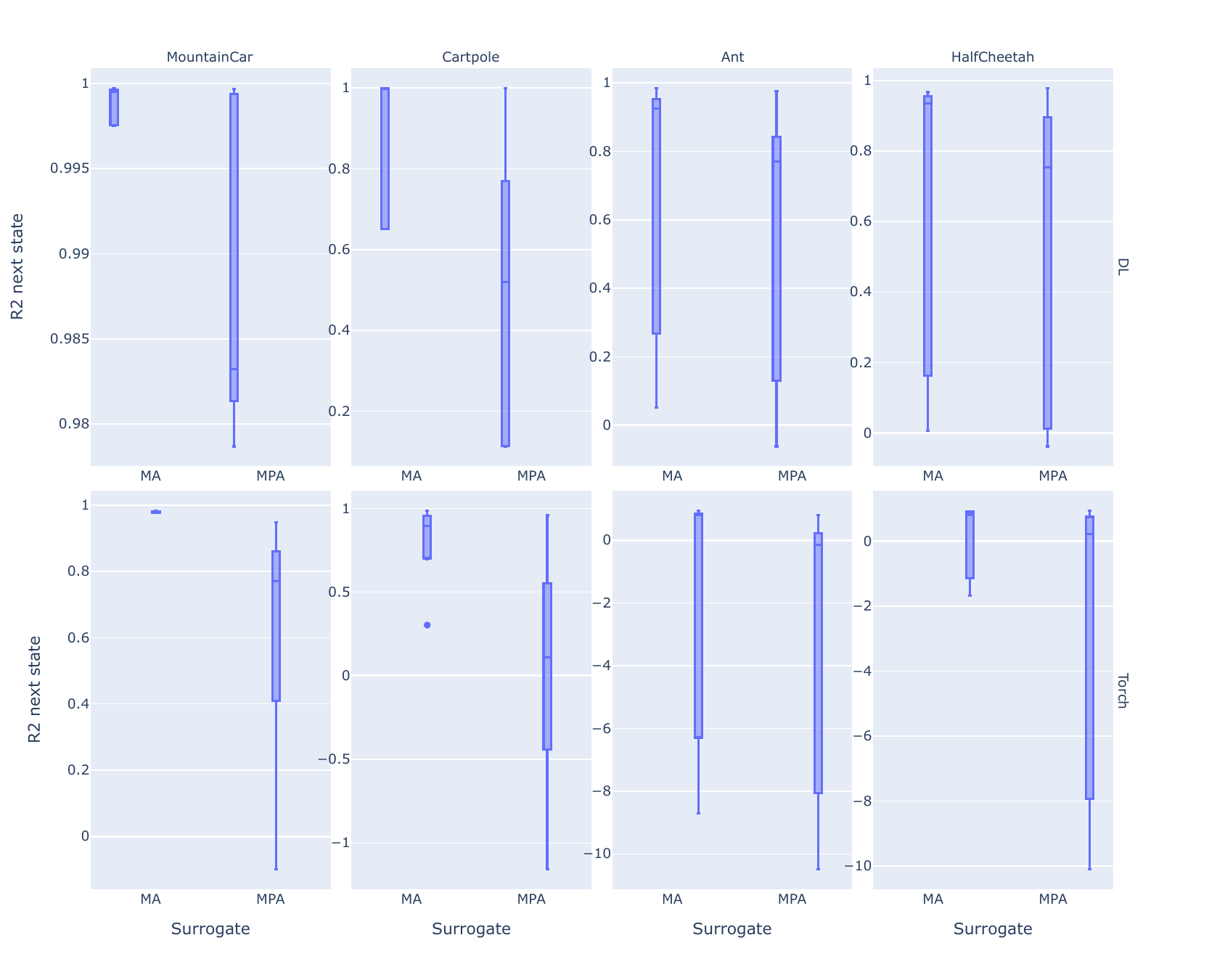}
   \caption{$R^2$ score distribution of \ac{MA} against \ac{MPA} over all the testing samplings for every environment. \ac{MA} shows better scores than \ac{MPA}, demonstrating the importance of the role of the \ac{MPA} dataset. This figure shows clearly that \ac{MA} beats \ac{MPA} in every scenario, which means that the effect of \ac{MEA} is fundamental for a good state space representation using RL agents as samplers.}
   \label{fig:ma_mpa}
\end{figure}

Finally, we compare specifically the effect of using the \ac{MEA} sampling method in the construction of a good surrogate capable of representing the overall dynamics of the environment. For that, we compare the difference between using MA, which mixes all the Agent-based sampling methods, with the \ac{MPA}, which excludes the \ac{MEA} method. Figure \ref{fig:ma_mpa} shows the distribution of both surrogates, and the picture clearly states that \ac{MA} is better than \ac{MPA}, so we can conclude from this that using \ac{MEA} is a significant improvement over using only some baseline agents like an expert agent, or some partially trained agent, or even using a random agent. This is due to the nature of the \ac{MEA} policy, which tries to maximize the entropy, (thus, the exploration), throughout all the variables of the environments. This is a guided process with a strategy to follow, so it can be expected to cover more dynamics of the environment than \ac{RA} or \ac{EA}, with the accuracy of having realistic trajectories of the environment and not samples of the whole space like the generative space samplers.

\section{Conclusions}

This paper presents a novel approach for constructing surrogate models using exploration techniques based on RL policies. The proposed method focuses on exploring realistic transitions in  simulated environments, resulting in high accuracy around the subset of states that appear in typical trajectories. We prove that using agent-based sampling methods is more effective than state-of-the-art generative methods, which evenly sample the state space of the simulated environment. This is further confirmed in environments with high dimensionality, while the improvement is less evident for smaller simulated environments. 

Surrogate modeling is often sample-inefficient. Depending on the speed of the simulators, the data acquisition phase may turn into an expensive process. In addition, it is possible that the accuracy presented in this work may not be enough for completely substituting real RL environments with these surrogates, as successive steps may compound the error, turning the visible environment into pure noise after few steps, as suggested by \cite{giraldo-perezReinforcementLearningBased2023}.

This work opens the door to challenges in the overlapping field of RL and surrogate models like using surrogate models for optimizing RL training processes. Furthermore, these results may also be extended to discrete scenarios, but this is left to prove.

As a brief summary, this work has shown the impact of using RL agents as samplers for building surrogate models, and we underline the critical effect of using agents that maximize the entropy of the simulated environment.

\section*{Funding}

This research has been supported by the Spanish Ministry (NextGenerationEU Funds) through Project IA4TES (Grant Number: MIA.2021.M04.0008).

%
%

%

\appendix
\label{appendix}
\section{Algorithm implementation details} \label{app:algo}

The XGBoost models are trained using the default parameters. For that, we use the XGBoost Python package \footnote{https://github.com/dmlc/xgboost}. 

The ANNs have been subjected to a hyperparameter optimization process. The main parameters are the following ones:
\begin{itemize}
    \item 2 hidden layers with 512 and 256 neurons each,
    \item learning rate of $0.001$,
    \item batch size of 64,
    \item 25 epochs for the Mujoco environments, and 10 epochs for the other environments.
    \item Early stopping if the validation curve was increased by more than $0.001$
\end{itemize}

The Gaussian surrogate (used in Kriging along with \ac{AL}) has the following specifications:
\begin{itemize}
    \item Uses \ac{LHS} for the general sampling procedure of each training step.
    \item Samples $100\,000$ space points every epoch.
    \item If the maximum std of the sampled space is less than $0.01$, the epoch is halted. Another stop condition is having added 300 points to the training set.
    \item We train 3 epochs. For each epoch, the initial space sample is repeated to prevent overfitting.
\end{itemize}

Note that the stopping conditions for the Kriging process have been added to prevent computational problems since, after every step, the training time increased drastically.

\end{document}